\pdfoutput=1

\documentclass[10pt,twocolumn,letterpaper]{article}

\usepackage{cvpr}
\usepackage{multirow}
\usepackage{times}
\usepackage{epsfig}
\usepackage{graphicx}
\usepackage{amsmath}
\usepackage{amssymb}
\usepackage{bm} 
\usepackage{tabulary}
\usepackage{booktabs}
\usepackage{lipsum}
\usepackage{hhline}
\usepackage[table,xcdraw]{xcolor}
\usepackage{enumitem}
\setitemize{noitemsep,topsep=5pt,parsep=5pt,partopsep=5pt}

\DeclareMathOperator*{\argmax}{argmax}
\DeclareMathOperator*{\softmax}{softmax}

\newcommand{\model}{\textsc{MotifNet}}
\newcommand{\modellong}{Stacked Motif Network}

\newcommand{\p}[1]{\textrm{Pr}( #1 )}  
\newcommand{\R}{\mathbb{R}}   
\newcommand{\card}[1]{\left\vert{#1}\right\vert}  
\newcommand{\bg}{\textsc{bg}}  
\newcommand{\term}[1]{\emph{#1}}  
\newcommand{\vect}[1]{\mathbf{#1}}   
\newcommand{\mat}[1]{\mathbf{#1}}    
\newcommand{\appropto}{\mathrel{\vcenter{
  \offinterlineskip\halign{\hfil$##$\cr
    \propto\cr\noalign{\kern2pt}\sim\cr\noalign{\kern-2pt}}}}}

\usepackage[pagebackref=true,breaklinks=true,letterpaper=true,colorlinks,bookmarks=false]{hyperref}

\cvprfinalcopy 

\def\cvprPaperID{4272} 

\begin{document}
\title{\vspace{-6mm}Neural Motifs: Scene Graph Parsing with Global Context}
\author{Rowan Zellers\textsuperscript{1}\quad Mark Yatskar\textsuperscript{1,2}\quad Sam Thomson\textsuperscript{3}\quad Yejin Choi\textsuperscript{1,2}
	\\ \textsuperscript{1}Paul G. Allen School of Computer Science \& Engineering, University of Washington
	\\ \textsuperscript{2}Allen Institute for Artificial Intelligence
	\\ \textsuperscript{3}School of Computer Science, Carnegie Mellon University \vspace{-1mm}
	\\ {\tt\small \{rowanz, my89, yejin\}@cs.washington.edu, sthomson@cs.cmu.edu}
	\\ \tt\normalsize \href{https://rowanzellers.com/neuralmotifs}{https://rowanzellers.com/neuralmotifs} \vspace{-5mm}
}


\maketitle

\begin{abstract}

We investigate the problem of producing structured graph representations of visual scenes. 
Our work analyzes the role of motifs: regularly appearing substructures in scene graphs.
We present new quantitative insights on such repeated structures in the Visual Genome dataset.
Our analysis shows that object labels are highly predictive of relation labels but not vice-versa.
We also find that there are recurring patterns even in larger subgraphs: more than 50\% of graphs contain motifs involving at least two relations.
Our analysis motivates a new baseline: given object detections, predict the most frequent relation between object pairs with the given labels, as seen in the training set.
This baseline improves on the previous state-of-the-art by an average of 3.6\% relative improvement across evaluation settings. 
We then introduce~{\modellong}s, a new architecture designed to capture higher order motifs in scene graphs that further improves over our strong baseline by an average 7.1\% relative gain.
Our code is available at \href{github.com/rowanz/neural-motifs}{github.com/rowanz/neural-motifs}.
\end{abstract}

\section{Introduction}
\begin{figure}
    \centering
    \includegraphics[scale=.23]{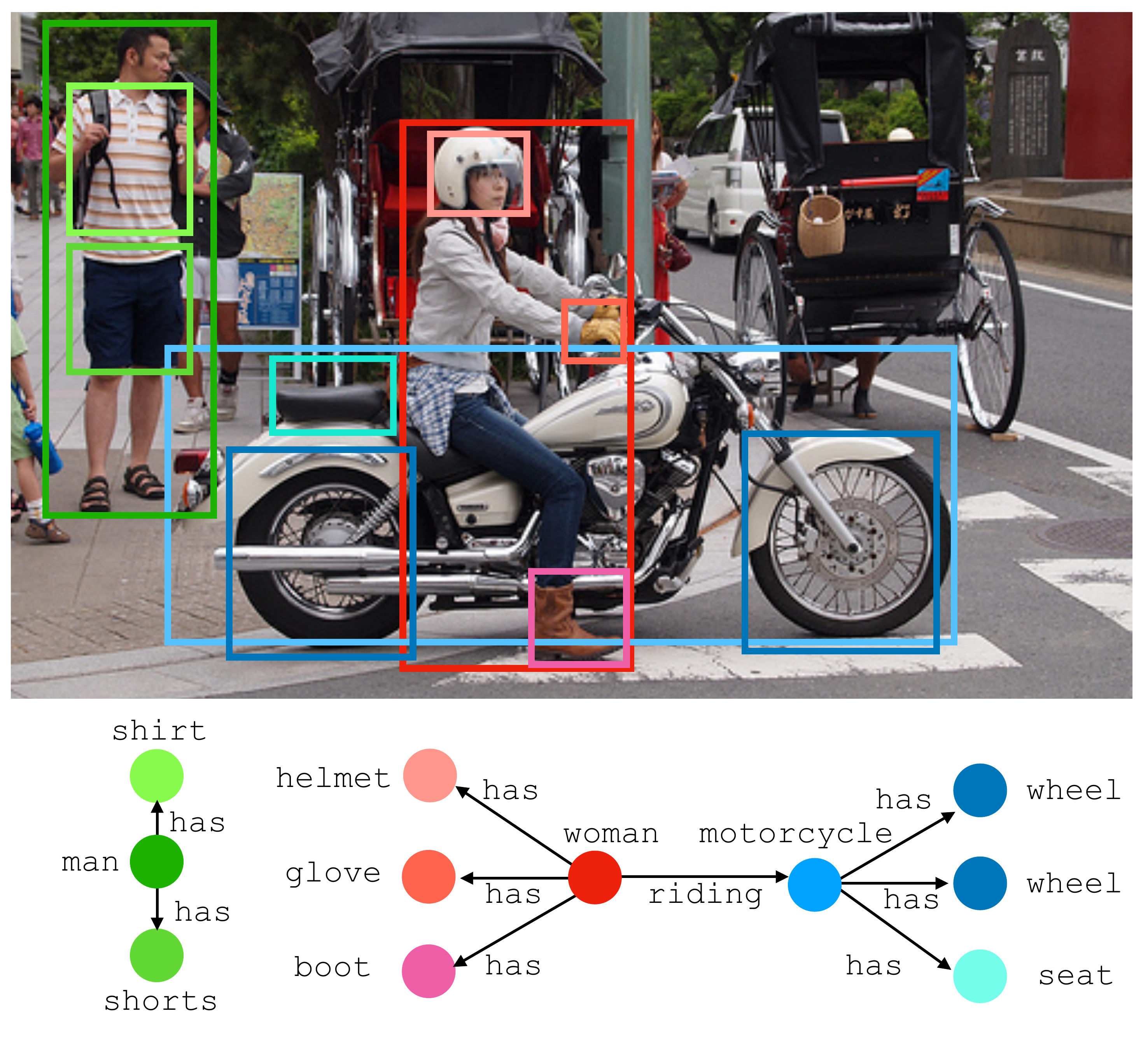}
    \caption{A ground truth scene graph containing entities, such as \texttt{woman}, \texttt{bike} or \texttt{helmet}, that are localized in the image with bounding boxes, color coded above, and the relationships between those entities, such as \texttt{riding}, the relation between \texttt{woman} and \texttt{motorcycle} or \texttt{has} the relation between \texttt{man} and \texttt{shirt}.
    }
    \label{fig:teaser}
\end{figure}

We investigate scene graph parsing: the task of producing graph representations of real-world images that provide semantic summaries of objects and their relationships.
For example, the graph in Figure~\ref{fig:teaser} encodes the existence of key objects such as people (``man'' and ``woman''), their possessions (``helmet'' and ``motorcycle'', both possessed by the woman), and their activities (the woman is ``riding'' the ``motorcycle'').
Predicting such graph representations has been shown to improve natural language based image tasks~\cite{johnson_image_2015,Teney2016GraphStructuredRF,Yin2017Obj2TextGV} 
and has the potential to significantly expand the scope of applications for computer vision systems.
Compared to object detection ~\cite{ren_faster_2015, redmon_yolo9000:_2016} , object interactions ~\cite{yao2010modeling,chao:iccv2015} and activity recognition ~\cite{2014survey}, 
scene graph parsing poses unique challenges since 
it requires reasoning about the complex dependencies across all of these components.

Elements of visual scenes have strong structural regularities.
For instance, people tend to wear clothes, as can be seen in Figure~\ref{fig:teaser}.
We examine these structural repetitions, or \emph{motifs}, using the Visual Genome~\cite{visualgenome} dataset, which provides annotated scene graphs for 100k images from COCO~\cite{mscoco}, consisting of over 1M instances of objects and 600k relations.
Our analysis leads to two key findings.
First, there are strong regularities in the local graph structure such that
the distribution of the relations is highly skewed once the corresponding object categories are given, but not vice versa.
Second, structural patterns exist even in larger
subgraphs; we find that over half of images contain previously occurring graph motifs. 

Based on our analysis, we introduce a simple yet powerful baseline: given object detections, predict the most frequent relation between object pairs with the given labels, as seen in the training set.
The baseline improves over prior state-of-the-art by 1.4 mean recall points (3.6\% relative), suggesting that an effective scene graph model must capture both the asymmetric dependence between objects and their relations, along with larger contextual patterns.

We introduce the \term{\modellong~(\model)}, a new neural network architecture
that complements existing approaches to scene graph parsing.
We posit that the key challenge in modeling scene graphs lies in devising an efficient mechanism to encode the global context that can directly inform the local predictors (i.e., objects and relations).
While previous work has used graph-based inference to propagate information in both directions between objects and relations~\cite{xu_scene_2017, li2017msdn, li_vip-cnn:_2017}, our analysis suggests strong independence assumptions in local predictors limit the quality of global predictions. 
Instead, our model predicts graph elements by staging bounding box predictions, object classifications, and relationships such that the global context encoding of all previous stages establishes rich context for predicting subsequent stages, as illustrated in Figure~\ref{fig:ourmodel}.
We represent 
the global context via recurrent sequential architectures such as Long Short-term Memory Networks (LSTMs) \cite{Hochreiter:1997:LSM:1246443.1246450}.

Our model builds on Faster-RCNN~\cite{ren_faster_2015}  for predicting bounding regions, fine tuned and adapted for Visual Genome.
Global context across bounding regions is computed and propagated through bidirectional LSTMs, which is then used by another LSTM that labels each bounding region conditioned on the overall context and all previous labels.
Another specialized layer of bidirectional LSTMs then computes and propagates information for predicting edges given bounding regions, their labels, and all other computed context.
Finally, we classify all $n^2$ edges in the graph, combining globally 
contextualized representations of head, tail, and image representations using using low-rank outer products~\cite{Kim2016HadamardPF}.
The method can be trained end-to-end.

Experiments on Visual Genome demonstrate the efficacy of our approach. First, we update existing work by pretraining the detector on Visual Genome, setting a new state-of-the-art (improving on average across evaluation settings 14.0 absolute points).
Our new simple baseline improves over previous work, using our updated detector, by a mean improvement of 1.4 points.
Finally, experiments show {\modellong}s is effective at modeling global context, with a mean improvement of 2.9 points (7.1\% relative improvement) over our new strong baseline. 

\label{sec:introduction}
\section{Formal definition}
\label{sec:definition}
\vspace{-5pt}
A \term{scene graph}, $G$, is a structured representation of the semantic content of an image~\cite{johnson_image_2015}.
It consists of:
\vspace*{-1mm}
\begin{itemize}
  \item a set $B = \{ b_1, \ldots, b_n \}$ of \term{bounding boxes}, 
  $b_i \in \mathbb{R}^4$,
  \vspace*{-1mm}
  \item a corresponding set $O = \{o_1, \ldots, o_n\}$ of \term{objects}, assigning a class label $o_i \in \mathcal{C}$ to each $b_i$, and
  \vspace*{-1mm}
  \item a set $R = \{r_1, \ldots, r_m\}$ of binary relationships between those objects.
\end{itemize}
\vspace*{-1mm}
Each relationship $r_k \in \mathcal{R}$ is a triplet of a start node $(b_i, o_i) \in B \times O$, an end node $(b_j, o_j) \in B \times O$, and a relationship label $x_{i \to j} \in \mathcal{R}$, where $\mathcal{R}$ is the set of all predicate types, including the ``background'' predicate, \bg, which indicates that there is no edge between the specified objects.
See Figure~\ref{fig:teaser} for an example scene graph.



\label{sec:problem}
\section{Scene graph analysis}
\label{sec:analysis}

\begin{table}[t]
    \small
    \centering
    \begin{tabular}{@{}c c c c@{}}
        \toprule

        Type & Examples & Classes & Instances \\

        \midrule

        \multicolumn{4}{c}{Entities}\\

        \cmidrule{1-4}

         Part & arm, tail, wheel & 32 & 200k (25.2\%)\\
         Artifact & basket, fork, towel & 34 & 126k (16.0\%) \\
         Person & boy, kid, woman & 13 & 113k (14.3\%) \\
         Clothes & cap, jean, sneaker & 16 & 91k (11.5\%) \\
         Vehicle & airplane, bike, truck,  & 12 &  44k (5.6\%)\\
         Flora & flower, plant, tree & 3 & 44k (5.5\%)\\
         Location & beach, room, sidewalk & 11 & 39k (4.9\%) \\
         Furniture & bed, desk, table & 9  & 37k (4.7\%)\\
         Animal & bear, giraffe, zebra  & 11 & 30k (3.8\%) \\
         Structure & fence, post, sign & 3 & 30k (3.8\%)\\
         Building & building, house & 2 & 24k (3.1\%)\\
         Food & banana, orange, pizza & 6  & 13k (1.6\%) \\

        \cmidrule{1-4}

        \multicolumn{4}{c}{Relations}\\

        \cmidrule{1-4}

         Geometric & above, behind, under & 15 & 228k (50.0\%)\\
         Possessive & has, part of, wearing & 8 & 186k (40.9\%)\\
         Semantic & carrying, eating, using & 24 & 39k (8.7\%)\\
         Misc & for, from, made of & 3 & 2k (0.3\%) \\

         \bottomrule
    \end{tabular}
    \caption{Object and relation types in Visual Genome, organized by super-type. Most, 25.2\% of entities are parts and 90.9\% of relations are geometric or possessive.}
    \label{tab:data_stat}
\end{table}

In this section, we seek quantitative insights on the structural regularities of scene graphs. In particular, (a) how different types of relations correlate with different objects, and (b) how higher order graph structures recur over different scenes. These insights motivate both the new baselines we introduce in this work and our model that better integrates the global context, described in  Section~\ref{sec:model}.

\subsection{Prevalent Relations in Visual Genome}
To gain insight into the Visual Genome scene graphs, we first categorize objects and relations into high-level types.
As shown in  Table~\ref{tab:data_stat}, the predominant relations are \term{geometric} and \term{possessive}, with clothing and parts making up over one third of entity instances.
Such relations are often obvious, e.g., houses tend to have windows.
In contrast, \term{semantic} relations, which correspond to activities, are less frequent and less obvious.
Although nearly half of relation types are semantic in nature, they comprise only 8.7\% of relation instances.
The relations ``using'' and ``holding''  account for 32.2\% of all semantic relation instances.

\begin{figure}[t]
    \vspace{-.2cm}
    \centering
    \includegraphics[scale=.23]{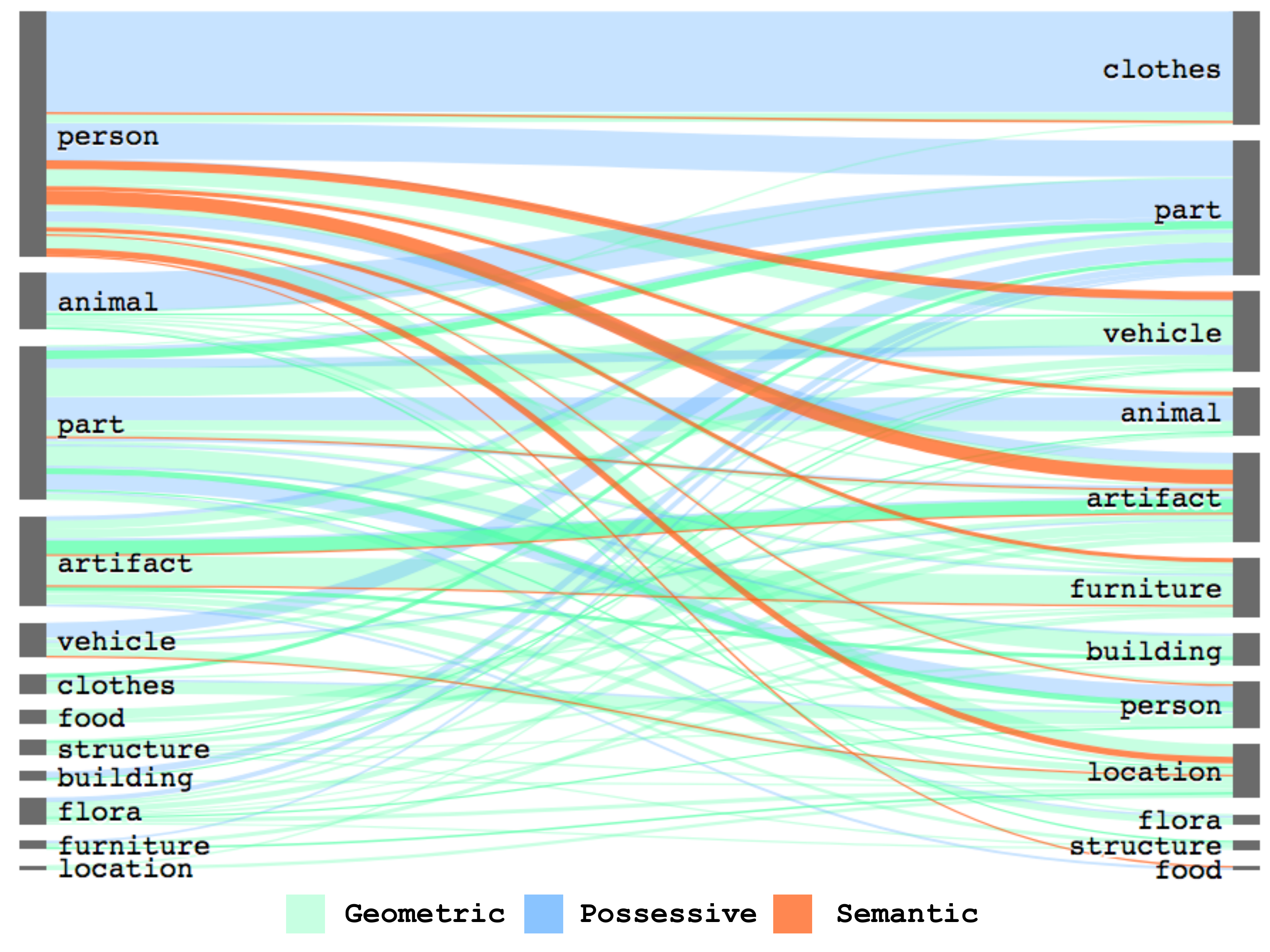}
    \caption{Types of edges between high-level categories in Visual Genome. Geometric, possessive and semantic edges cover 50.9\%, 40.9\%, and 8.7\%, respectively, of edge instances in scene graphs. The majority of semantic edges occur between people and vehicles, artifacts and locations. 
    Less than 2\% of edges between clothes and people are semantic.}
    \label{fig:edges}
\end{figure}

Using our high-level types, we visualize the distribution of relation types between object types in Figure~\ref{fig:edges}.
Clothing and part entities are almost exclusively linked through possessive relations while furniture and building entities are almost exclusively linked through geometric relations.
Geometric and spatial relationships between certain entities are interchangeable, for example,
when a ``part'' is the head object, it tends to connect to other entities through a geometric relation (e.g. wheel on bike); when a ``part'' is the tail object, it tends to be connected with possessive relations (e.g. bike has wheel).
Nearly all semantic relationship are headed by people, with the majority of edges relating to artifacts, vehicles, and locations.
Such structural predictability and the prevalence of geometric and part-object relations suggest that common sense priors play an important role in generating accurate scene graphs.

\begin{figure}[t]
    \vspace{-.3cm}
    \centering
    \includegraphics[scale=.23]{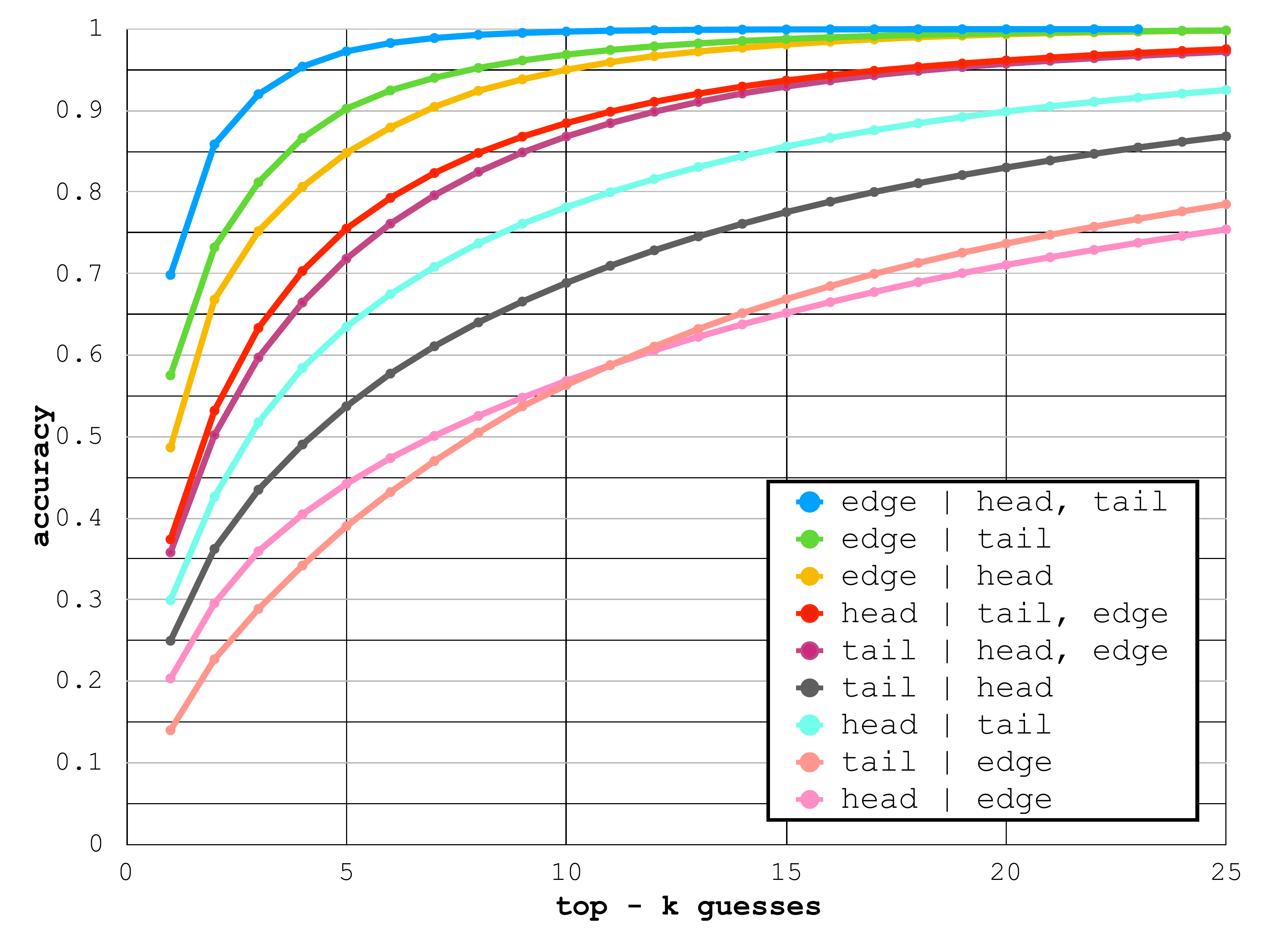}
    \caption{The likelihood of guessing, in the top-k, head, tail, or edge labels in a scene graph, given other graph components (i.e. without image features). Neither head nor tail labels are strongly determined by other labels, but given the identity of head and tail, edges (\texttt{edge} $|$ \texttt{head, tail}) can be determined with 97\% accuracy in under 5 guesses. Such strong biases make it critical to condition on objects when predicting edges.} 
    \label{fig:conditionals}
\end{figure}

In Figure~\ref{fig:conditionals}, we examine how much information is gained by knowing the identity of different parts 
in a scene graphs.
In particular, we consider how many guesses are required to determine the labels of head (h), edge (e) or tail (t) given labels of the other elements, only using label statistics computed on scene graphs.
Higher curves imply that the element is highly determined given the other values.
The graph shows that the local distribution of relationships has significant structure.
In general, the identity of edges involved in a relationship is not highly informative of other elements of the structure while the identities of head or tail provide significant information, both to each other and to edge labels.
Adding edge information to already given head or tail information provides minimal gain.
Finally, the graph shows edge labels are highly determined given the identity of object pairs:
the most frequent relation is correct 70\% of the time, and
the five most frequent relations for the pair contain the correct label 97\% of the time.

\subsection{Larger Motifs}

\begin{figure}[t]
    \vspace{-.3cm}
    \centering
    \includegraphics[scale=.23]{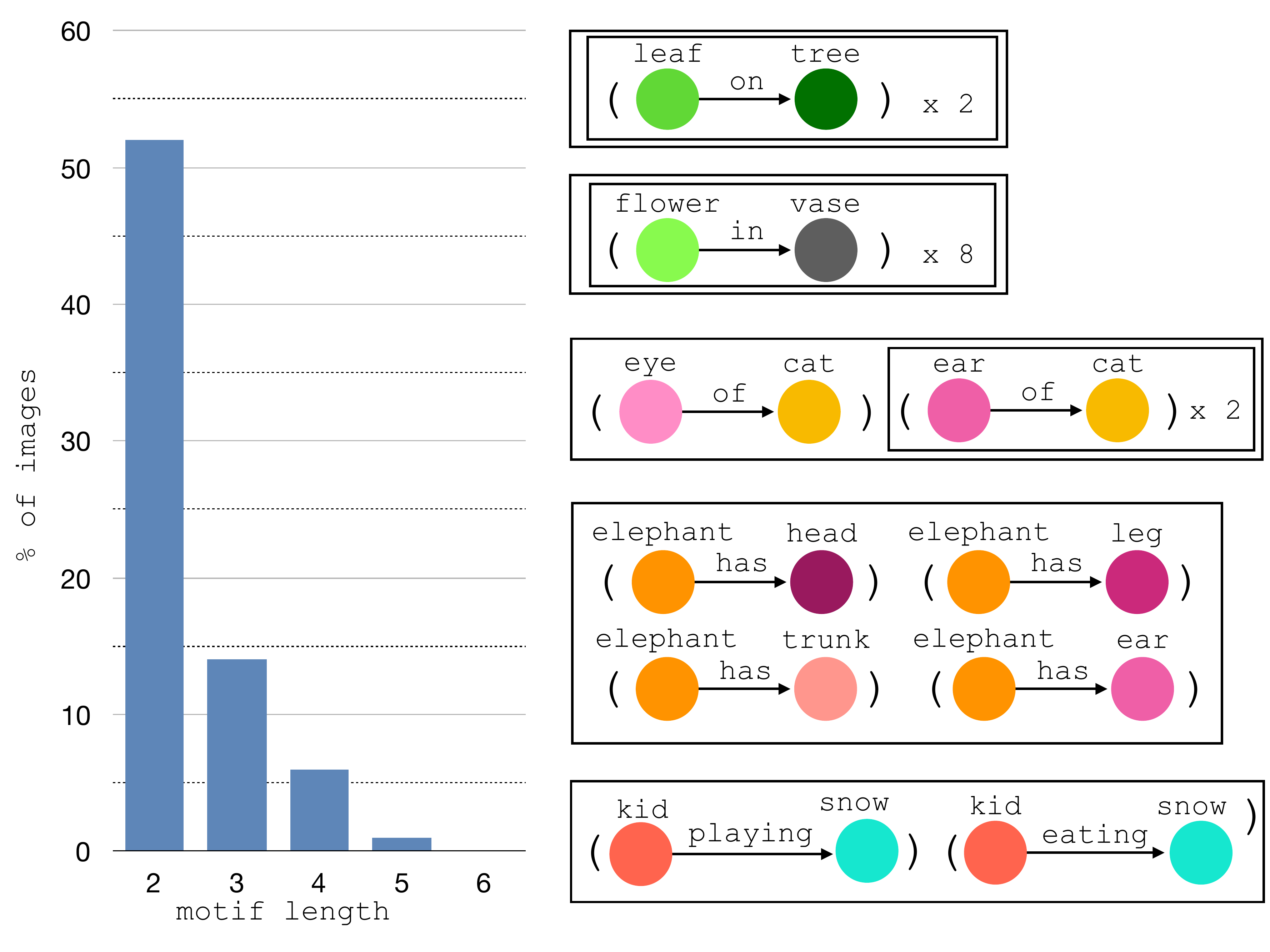}
    \caption{On the left, the percent of images that have a graph motif found in Visual Genome using pointwise mutual information, composed of at least a certain length (the number of edges it contains). Over 50\% of images have at least one motif involving two relationships. On the right, example motifs, where structures repeating many times is indicated with plate notation. For example, the second motif is length 8 and consists of 8 flower-in-vase relationships. Graph motifs commonly result from groups (e.g., several instances of ``leaf on tree''), and correlation between parts (e.g., ``elephant has head,'' ``leg,'' ``trunk,'' and ``ear.''). }
    \label{fig:motifs}
\end{figure}

Scene graphs not only have local structure but have higher order structure as well.
We conducted an analysis of repeated motifs in scene graphs by mining combinations of object-relation-object labels that have high pointwise mutual information with each other.
Motifs were extracted iteratively: first we extracted motifs of two combinations, replaced all instances of that motif with an atomic symbol and mined new motifs given previously identified motifs.
Combinations of graph elements were selected as motifs if both elements involved occurred at least 50 times in the Visual Genome training set and were at least 10 times more likely to occur together than apart.
Motifs were mined until no new motifs were extracted.
Figure~\ref{fig:motifs} contains example motifs we extracted on the right, and the prevalence of motifs of different lengths in images on the left.
Many motifs correspond to either combinations of parts, or objects that are commonly grouped together.
Over 50\% of images in Visual Genome contain a motif involving at least two combinations of object-relation-object, and some images contain motifs involving as many as 16 elements.

\section{Model}
\label{sec:model}

\begin{figure*}[t]
    \vspace{-3mm}
    \centering
    \includegraphics[scale=.63]{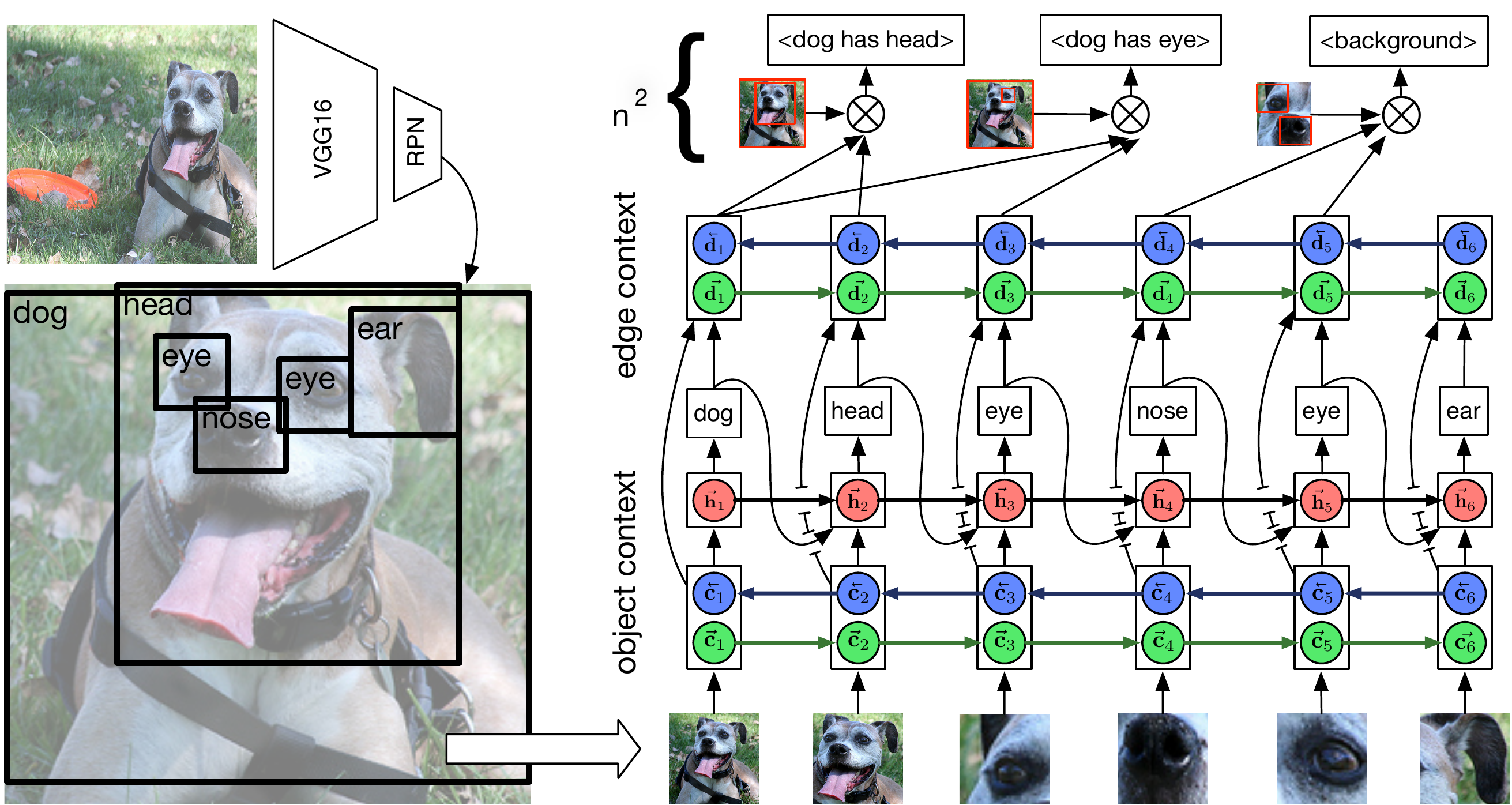}
    \caption{A diagram of a \modellong~ (\model). The model breaks scene graph parsing into stages predicting bounding regions, labels for regions, and then relationships. Between each stage, global context is computed using bidirectional LSTMs and is then used for subsequent stages. In the first stage, a detector proposes bounding regions and then contextual information among bounding regions is computed and propagated (object context). The global context is used to predict labels for bounding boxes. Given bounding boxes and labels, the model constructs a new representation (edge context) that gives global context for edge predictions. Finally, edges are assigned labels by combining contextualized head, tail, and union bounding region information with an outer product.\vspace{-1mm}}
    \label{fig:ourmodel}

\end{figure*}

Here we present our novel model, \term{\modellong}~(\model).
\model\ decomposes the probability of a graph $G$ (made up of a set of bounding regions $B$, object labels $O$, and labeled relations $R$) into three factors:
\begin{equation}
\p{G \mid I} = \p{B \mid I} \ \p{O \mid B, I} \ \p{R \mid B, O, I}.
\end{equation}
Note that this factorization makes no independence assumptions.
Importantly, predicted object labels may depend on one another, and predicted relation labels may depend on predicted object labels.
The analyses in Section~\ref{sec:analysis} make it clear that capturing these dependencies is crucial.

The \emph{bounding box} model ($\p{B \mid I}$) is a fairly standard object detection model, which we describe in Section~\ref{subsec:model:bounding_boxes}.
The \emph{object} model ($\p{O \mid B, I}$; Section~\ref{subsec:model:objects}) conditions on a potentially large set of predicted bounding boxes, $B$.
To do so, we linearize $B$ into a sequence that an LSTM then processes to create a contextualized representation of each box.
Likewise, when modeling \emph{relations} ($\p{R \mid B, O, I}$; Section~\ref{subsec:model:relations}), we linearize the set of predicted labeled objects, $O$, and process them with another LSTM to create a representation of each object in context.
Figure~\ref{fig:ourmodel} contains a visual summary of the entire model architecture.

\subsection{Bounding Boxes}
\label{subsec:model:bounding_boxes}

We use Faster R-CNN as an underlying detector~\cite{ren_faster_2015}.
For each image $I$, the detector predicts a set of region proposals $B = \{ b_1, \ldots, b_n \}$.
For each proposal $b_i \in B$ it also outputs a feature vector $\vect{f}_i$ and a vector $\vect{l}_i \in \R^{\card{\mathcal{C}}}$ of (non-contextualized) object label probabilities.
Note that because \bg\ is a possible label, our model has not yet committed to any bounding boxes.
See Section~\ref{subsec:setup:model_details} for details.

\subsection{Objects}
\label{subsec:model:objects}

\paragraph{Context}
We construct a contextualized representation for object prediction based on the set of proposal regions $B$.
Elements of $B$ are first organized into a linear sequence, $[(b_1,\vect{f}_1,\vect{l}_1), \ldots, (b_n,\vect{f}_n, \vect{l}_n)]$.\footnote{We consider several strategies to order the regions, see Section ~\ref{subsec:setup:model_details}.}
The \term{object context}, $\mat{C}$, is then computed using a bidirectional LSTM~\cite{Hochreiter:1997:LSM:1246443.1246450}:
\begin{equation}
\mat{C} = \text{biLSTM}(
  [\vect{f}_i; \mat{W}_1\vect{l}_i]_{i=1,\ldots,n}
),
\end{equation}
$\mat{C} = [\vect{c}_1, \ldots, \vect{c}_n]$ contains the final LSTM layer's hidden states for each element in the linearization of $B$, and $\mathbf{W}_1$ is a parameter matrix that maps the distribution of predicted classes, $\vect{l}_1$, to $\R^{100}$. The biLSTM allows all elements of $B$ to contribute information about potential object identities.

\paragraph{Decoding}
The context $\mat{C}$ is used to sequentially decode labels for each proposal bounding region, conditioning on previously decoded labels.
We use an LSTM to decode a category label for each contextualized representation in $\mat{C}$:
\begin{align}
    \vect{h}_i &= \text{LSTM}_i\left( [\vect{c}_{i}; \vect{\hat{o}}_{i-1}] \right)\\
    \label{eqn:max_o}
    \vect{\hat{o}}_i &= \argmax{\left( \mat{W}_o~\vect{h}_i \right)} \in \R^{\card{\mathcal{C}}} \text{ (one-hot)}
\end{align}
We then discard the hidden states $\vect{h}_i$ and use the object class commitments $\vect{\hat{o}}_i$ in the relation model (Section \ref{subsec:model:relations}).

\subsection{Relations}
\label{subsec:model:relations}

\paragraph{Context}
We construct a contextualized representation of bounding regions, $B$, and objects, $O$, using additional bidirectional LSTM layers:
\begin{equation}
\mat{D} = \text{biLSTM}(
  [\vect{c}_i; \mat{W}_2 \vect{\hat{o}}_i]_{i=1,\ldots,n}
),
\end{equation}
where the \term{edge context}
$\mat{D} = [\vect{d}_{1}, \ldots, \vect{d}_{n}]$ contains the states for each bounding region at the final layer, and $\mat{W}_2$ is a parameter matrix mapping $\vect{\hat{o}}_i$ into $\R^{100}$.

\paragraph{Decoding}
There are a quadratic number of possible relations in a scene graph.
For each possible edge, say between $b_i$ and $b_j$, we compute the probability the edge will have label $x_{i \to j}$ (including \bg).
The distribution uses global context, $\mat{D}$, and a feature vector for the union of boxes~\footnote{A union box is the convex hull of the union of two bounding boxes.}, $\vect{f}_{i,j}$:
\begin{align}
\label{eq:hadamard}
      \vect{g}_{i, j} &= (\mat{W}_h \vect{d}_{i}) \circ (\mat{W}_t \vect{d}_{j}) \circ \vect{f}_{i, j} \\
      \p{ x_{i \to j} \mid B, O } &=
 \softmax{\left(
   \mat{W}_r \vect{g}_{i, j} + \vect{w}_{o_i,o_j}
 \right)}.
\end{align}
$\mat{W}_h$ and $\mat{W}_t$ project the head and tail context into $\R^{4096}$.
$\vect{w}_{o_i,o_j}$ is a bias vector specific to the head and tail labels.

\section{Experimental Setup}
\label{sec:setup}
In the following sections we explain (1) details of how we construct the detector, order bounding regions, and implement the final edge classifier (Section~\ref{subsec:setup:model_details}), (2) details of training (Section~\ref{subsec:setup:training}), and (3) evaluation (Section~\ref{subsec:setup:evaluation}).

\subsection{Model Details}
\label{subsec:setup:model_details}
\paragraph{Detectors}
Similar to prior work in scene graph parsing~\cite{xu_scene_2017, li2017msdn}, we use Faster RCNN with a VGG backbone as our underling object detector \cite{ren_faster_2015,Simonyan14c}. Our detector is given images that are scaled and then zero-padded to be 592x592. We adjust the bounding box proposal scales and dimension ratios to account for different box shapes in Visual Genome, similar to YOLO-9000~\cite{redmon_yolo9000:_2016}. To control for detector performance in evaluating different scene graph models, we first pretrain the detector on Visual Genome objects. We optimize the detector using SGD with momentum on 3 Titan Xs, with a batch size of $b=18$, and a learning rate of $lr=1.8\cdot 10^{-2}$ that is divided by 10 after validation mAP plateaus.
For each batch we sample 256 RoIs per image, of which 75\% are background. The detector gets 20.0 mAP (at 50\% IoU) on Visual Genome; the same model, but trained and evaluated on COCO, gets 47.7 mAP at 50\% IoU. Following \cite{xu_scene_2017}, we integrate the use the detector freezing the convolution layers and duplicating the fully connected layers, resulting in separate branches for object/edge features.
\paragraph{Alternating Highway LSTMs}
To mitigate vanishing gradient problems as information flows upward, we add highway connections to all LSTMs \cite{he2017deep, Srivastava:2015:TVD:2969442.2969505, zhang_highway_lstm}. 
To additionally reduce the number of parameters, we follow \cite{he2017deep} and alternate the LSTM directions. Each alternating highway LSTM step can be written as the following wrapper around the conventional LSTM equations \cite{Hochreiter:1997:LSM:1246443.1246450}:
\begin{align}
\mathbf{r}_i &= \sigma(\mathbf{W}_g[\mathbf{h}_{i-\delta}, \mathbf{x}_i] + \mathbf{b}_g) \\
\mathbf{h}_i &= \mathbf{r_i} \circ \text{LSTM}(\mathbf{x}_i, \mathbf{h}_{i-\delta}) + (1-\mathbf{r_i}) \circ \mathbf{W}_i\mathbf{x}_i,
\end{align}
where $\mathbf{x}_i$ is the input, $\mathbf{h}_i$ represents the hidden state, and $\delta$ is the direction: $\delta=1$ if the current layer is even, and $-1$ otherwise. For \model, we use 2 alternating highway LSTM layers for object context, and 4 for edge context.
\paragraph{RoI Ordering for LSTMs}
We consider several ways of ordering the bounding regions:
\begin{enumerate}[label={(\arabic*)},noitemsep,nolistsep]
  \setlength{\parskip}{0pt}
  \setlength{\parsep}{0pt}
    \item \textsc{LeftRight} (default): Our default option is to sort the regions left-to-right by the central x-coordinate: we expect this to encourage the model to predict edges between nearby objects, which is beneficial as objects appearing in relationships tend to be close together.
    \item \textsc{Confidence}: Another option is to order bounding regions based on the confidence of the maximum non-background prediction from the detector: $\max_{j \ne \bg} \vect{l}_i^{(j)}$, as this lets the detector commit to ``easy'' regions, obtaining context for more difficult regions.\footnote{When sorting by confidence, the edge layer's regions are ordered by the maximum non-background object prediction as given by Equation~\ref{eqn:max_o}.}
    \item \textsc{Size}: Here, we sort the boxes in descending order by size, possibly predicting global scene information first.
    \item \textsc{Random}: Here, we randomly order the regions.
\end{enumerate}
\paragraph{Predicate Visual Features} To extract visual features for a predicate between boxes $b_i, b_j$, we resize the detector's features corresponding to the union box of $b_i,b_j$ to 7x7x256. We model geometric relations using a 14x14x2 binary input with one channel per box. We apply two convolution layers to this and add the resulting 7x7x256 representation to the detector features. Last, we apply finetuned VGG fully connected layers to obtain a 4096 dimensional representation.\footnote{We remove the final ReLU to allow more interaction in Equation~\ref{eq:hadamard}.}
\subsection{Training}
\label{subsec:setup:training}
We train~\model~on ground truth boxes, with the objective to predict object labels and to predict edge labels given ground truth object labels. For an image, we include all annotated relationships (sampling if more than 64) and sample 3 negative relationships per positive.
In cases with multiple edge labels per directed edge (5\% of edges), we sample the predicates. Our loss is the sum of the cross entropy for predicates and cross entropy for objects predicted by the object context layer. We optimize using SGD with momentum on a single GPU, with $lr=6\cdot10^{-3}$ and $b=6$.

\paragraph{Adapting to Detection}
Using the above protocol gets good results when evaluated on scene graph classification, but models that incorporate context underperform when suddenly introduced to non-gold proposal boxes at test time.

To alleviate this, we fine-tune using noisy box proposals from the detector.
We use per-class non-maximal suppression (NMS) \cite{NMSCITATION} at 0.3 IoU to pass 64 proposals to the object context branch of our model. We also enforce NMS constraints during decoding given object context.
We then sample relationships between proposals that intersect with ground truth boxes and use relationships involving these boxes to finetune the model until detection convergence.

We also observe that in detection our model gets swamped with many low-quality RoI pairs as possible relationships, which slows the model and makes training less stable. To alleviate this, we observe that nearly all annotated relationships are between overlapping boxes,\footnote{A hypothetical model that perfectly classifies relationships, but only between boxes with nonzero IoU, gets 91\% recall.} and classify all relationships with non-overlapping boxes as \bg.

\subsection{Evaluation}
\label{subsec:setup:evaluation}
We train and evaluate our models on Visual Genome, using the publicly released preprocessed data and splits from \cite{xu_scene_2017}, containing 150 object classes and 50 relation classes, but sample a development set from the training set of 5000 images. 
We follow three standard evaluation modes: (1) \textbf{predicate classification} (\textsc{PredCls}): given a ground truth set of boxes and labels, predict edge labels, (2) \textbf{scene graph classification} (\textsc{SGCls}): given ground truth boxes, predict box labels and edge label and (3) \textbf{scene graph detection} (\textsc{SGDet}): predict boxes, box labels, and edge labels.
The annotated graphs are known to be incomplete,
thus systems are evaluated using recall@$K$ metrics.\footnote{Past work has considered these evaluation modes at recall thresholds R$@50$ and R$@100$, but we also report results on R$@20$.}

In all three modes, recall is calculated for relations; a ground truth edge $(b_h, o_h, x, b_t, o_t)$ is counted as a ``match'' if there exist predicted boxes $i,j$ such that $b_i$ and $b_j$ respectively have sufficient overlap with $b_h$ and $b_t$,\footnote{As in prior work, we compute the intersection-over-union (IoU) between the boxes and use a threshold of 0.5.} and the objects and relation labels agree.
We follow previous work in enforcing that for a given head and tail bounding box, the system must not output multiple edge labels \cite{xu_scene_2017, lu_visual_2016}.

\subsection{Frequency Baselines}
To support our finding that object labels are highly predictive of edge labels, we additionally introduce several frequency baselines built off training set statistics. The first, \textsc{Freq}, uses our pretrained detector to predict object labels for each RoI. To obtain predicate probabilities between boxes $i$ and $j$, we look up the empirical distribution over relationships between objects $o_i$ and $o_j$ as computed in the training set.\footnote{Since we consider an edge $x_{i \to j}$ to have label \bg\ if $o$ has no edge to $j$, this gives us a valid probability distribution.} Intuitively, while this baseline does not look at the image to compute $\textrm{Pr}(x_{i\to j} | o_i, o_j)$, it displays the value of \emph{conditioning} on object label predictions $o$. The second, \textsc{Freq-Overlap}, requires that the two boxes intersect in order for the pair to count as a valid relation.

\section{Results}
\label{sec:results}
\begin{table*}[htbp]
\vspace{-.2cm}
\small
\centering
\begin{tabular}{@{}c@{\hspace{0.4em}} l c@{\hspace{0.2em}} ccc c@{\hspace{0.2em}} ccc c@{\hspace{0.2em}} ccc c@{\hspace{0.2em}} c@{}}
\toprule
      && \phantom{} & \multicolumn{3}{c}{Scene Graph Detection} &  \phantom{} & \multicolumn{3}{c}{Scene Graph Classification} &  \phantom{} & \multicolumn{3}{c}{Predicate Classification} & \phantom{} & Mean\\
    \cmidrule{4-6} \cmidrule{8-10} \cmidrule{12-14} 
& Model && R$@$20 & R$@$50  & R$@$100 && R$@$20 & R$@$50  & R$@$100  && R$@$20 & R$@$50  & R$@$100 &&  \\
\midrule
\multirow{7}{*}{\rotatebox[origin=c]{90}{models}} & \sc Vrd \cite{lu_visual_2016} && & 0.3  & 0.5 && & 11.8  & 14.1  && & 27.9  & 35.0 && 14.9\\
&\sc Message Passing \cite{xu_scene_2017}&&  & 3.4  & 4.2 &&  &21.7 & 24.4 && & 44.8 & 53.0 && 25.3 \\
&\sc Message Passing$+$ && 14.6 & 20.7 & 24.5 && 31.7 & 34.6 & 35.4 && 52.7 & 59.3 & 61.3 && 39.3 \\
&\sc Assoc Embed \cite{DBLP:journals/corr/NewellD17}$\star$ && 6.5 & 8.1 & 8.2 && 18.2 & 21.8 & 22.6 && 47.9 & 54.1 & 55.4 && 28.3 \\ 
& \textsc{Freq} && 17.7 & 23.5 & 27.6 && 27.7 & 32.4 & 34.0 && 49.4 & 59.9 & 64.1 && 40.2 \\
&\textsc{Freq+Overlap} && 20.1 & 26.2 & 30.1 && 29.3 & 32.3 & 32.9 && 53.6 & 60.6 & 62.2 && 40.7 \\
& \model-\textsc{LeftRight} &&21.4 & 27.2 & 30.3 && {\bf 32.9} & {\bf 35.8} & {\bf 36.5} && {\bf 58.5} & {\bf 65.2} & {\bf 67.1} && {\bf 43.6}\\ \hline
\multirow{4}{*}{\rotatebox[origin=c]{90}{ablations}}
&\model-\textsc{NoContext} && 21.0 & 26.2 & 29.0 && 31.9 & 34.8 & 35.5 && 57.0 & 63.7 & 65.6 && 42.4\\ 
& 
\model-\textsc{Confidence} &&{\bf 21.7} & {\bf 27.3} & {\bf 30.5} && 32.6 & 35.4 & 36.1 && 58.2 & 65.1 & 67.0 && 43.5 \\
&\model-\textsc{Size} && 21.6 & {\bf 27.3} & 30.4 && 32.2 & 35.0 & 35.7 && 58.0 & 64.9 & 66.8 && 43.3\\
&\model-\textsc{Random} && 21.6 & {\bf 27.3} & 30.4 && 32.5 & 35.5 & 36.2 && 58.1 & 65.1 & 66.9 && 43.5\\ 
\bottomrule

\end{tabular}
\label{tab:superresults}
\caption{Results table, adapted from \cite{xu_scene_2017} which ran VRD \cite{lu_visual_2016} without language priors. All numbers in \%. Since past work doesn't evaluate on R$@$20, we compute the mean by averaging performance on the 3 evaluation modes over R$@$50 and R$@$100. $\star$: results in \cite{DBLP:journals/corr/NewellD17} are without scene graph constraints; we evaluated performance with constraints using saved predictions given to us by the authors 
(see Table~\ref{tab:supptable} in supp).
} 
\vspace{-0.1cm}
\end{table*}

We present our results in Table~\ref{tab:superresults}.
We compare \model~to previous models not directly incorporating context (\textsc{Vrd} \cite{lu_visual_2016} and \textsc{Assoc Embed} \cite{DBLP:journals/corr/NewellD17}), a state-of-the-art approach for incorporating graph context via message passing (\textsc{Message Passing}) \cite{xu_scene_2017}, and its re-implemenation using our detector, edge model, and NMS settings (\textsc{Message Passing+}). Unfortunately, many scene graph models are evaluated on different versions of Visual Genome; see 
Table~\ref{tab:supptable} in 
the supp for more analysis. 

Our best frequency baseline, \textsc{Freq+Overlap}, improves over prior state-of-the-art by 1.4 mean recall, primarily due to improvements in detection and predicate classification, where it outperforms \textsc{Message Passing+} by 5.5 and 6.5 mean points respectively. 
\model~improves even further, by 2.9 additional mean points over the baseline (a 7.1\% relative gain).  


\paragraph{Ablations} To evaluate the effectiveness of our main model,~\model,~we consider several ablations in Table~\ref{tab:superresults}. In \model-\textsc{NoContext}, we predict objects based on the fixed detector, and feed non-contexualized embeddings of the head and tail label into Equation~\ref{eq:hadamard}. Our results suggest that there is signal in the vision features for edge predictions, as \model-\textsc{NoContext} improves over \textsc{Freq-Overlap}. Incorporating context is also important: 
our full model~\model~improves by 1.2 mean points, with largest gains at the lowest recall threshold of R$@$20. \footnote{The larger improvement at the lower thresholds suggests that our models mostly improve on relationship ordering rather than classification. Indeed, it is often unnecessary to order relationships at the higher thresholds: 51\% of images have fewer than 50 candidates and 78\% have less than 100.}
We additionally validate the impact of the ordering method used, as discussed in Section~\ref{subsec:setup:model_details}; the results vary less than 0.3 recall points, suggesting that~\model~is robust to the RoI ordering scheme used.

\section{Qualitative Results}
\label{sec:qualitative}
\begin{figure*}[t]
    \vspace{-.3cm}
    \centering
    \includegraphics[scale=0.51]{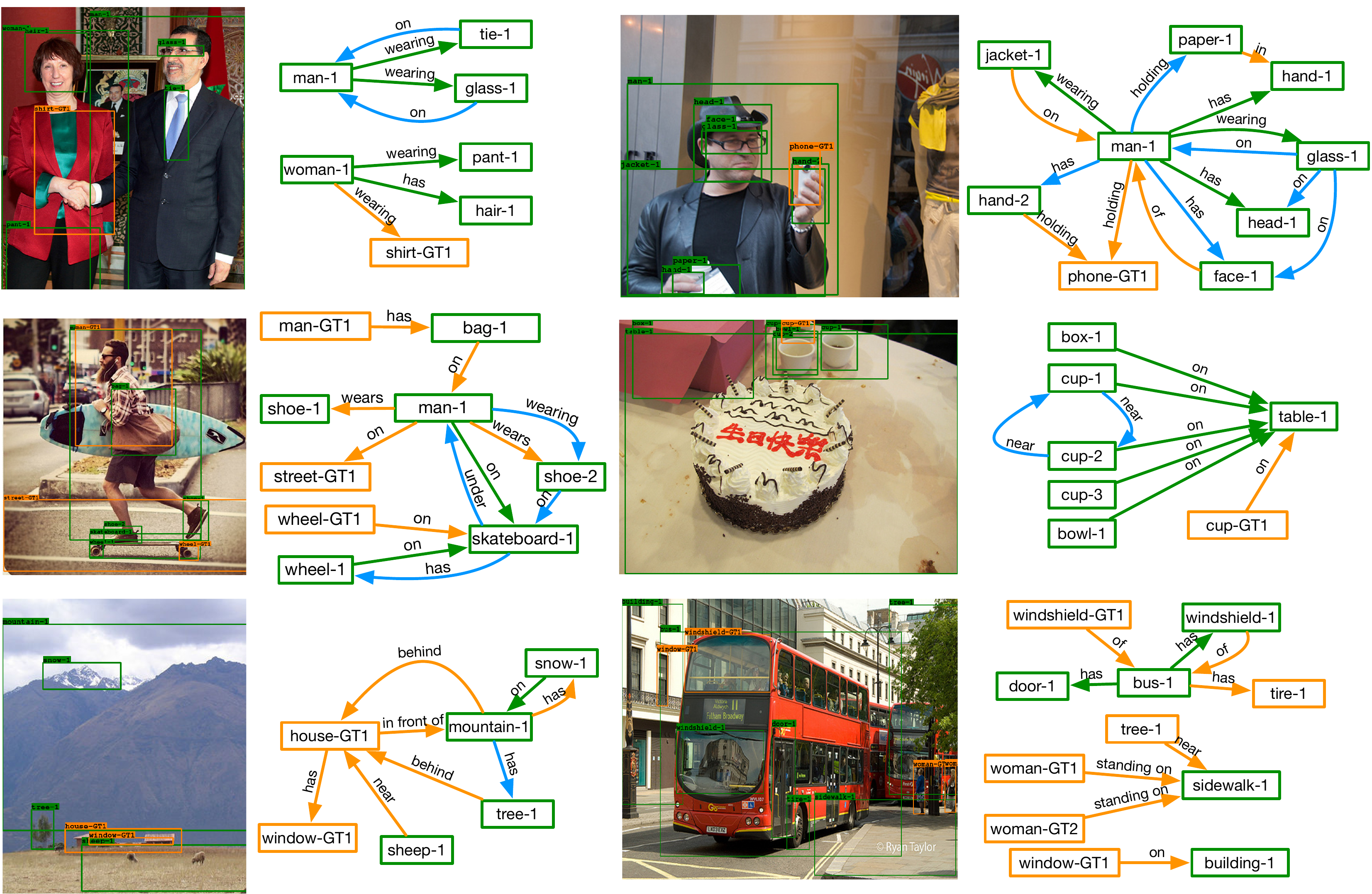}
    \caption{Qualitative examples from our model in the Scene Graph Detection setting. Green boxes are predicted and overlap with the ground truth, orange boxes are ground truth with no match. Green edges are true positives predicted by our model at the R@20 setting, orange edges are false negatives, and blue edges are false positives. Only predicted boxes that overlap with the ground truth are shown. }
    \vspace{-.4cm}
    \label{fig:examples}
\end{figure*}

Qualitative examples of our approach, shown in Figure~\ref{fig:examples}, suggest that~\model~ is able to induce graph motifs from detection context. Visual inspection of the results suggests that the method works even better than the quantitative results would imply, since many seemingly correct edges are predicted that do not exist in the ground truth.

There are two common failure cases of our model. The first, as exhibited by the middle left image in Figure~\ref{fig:examples} of a skateboarder carrying a surfboard, stems from predicate ambiguity (``wearing'' vs ``wears'').
The second common failure case occurs when the detector fails, resulting in cascading failure to predict any edges to that object.
For example, the failure to predict ``house'' in the lower left image resulted in five false negative relations.

\section{Related Work}
\label{sec:relation_work}
\paragraph{Context}
Many methods have been proposed for modeling semantic context in object recognition~\cite{divvala2009empirical}. 
Our approach is most closely related to work that models object co-occurrence using graphical models to combine many sources of contextual information~\cite{rabinovich2007objects,galleguillos2010context,li2007,farhadi2010every}.
While our approach is a type of graphical model, it is unique in that it stages incorporation of context allowing for meaningful global context from large conditioning sets.

Actions and relations have been a particularly fruitful source of context~\cite{marszalek2009actions, yatskar_situation_2016}, especially when combined with pose to create human-object interactions~\cite{yao2010modeling,chao:iccv2015}. Recent work has shown that object layouts can provide sufficient context for captioning COCO images~\cite{obj2textEMNLP2017,mscoco}; our work suggests the same for parsing Visual Genome scene graphs.
Much of the context we derive could be interpreted as commonsense priors, which have commonly been extracted using auxiliary means~\cite{zhu2014reasoning, viske, neil, Yatskar_VCommonSense_16, emnlp17_zellers}. 
Yet for scene graphs, we are able to directly extract such knowledge.

\paragraph{Structured Models}
Structured models in visual understanding have been explored for language grounding, where language determines the graph structures involved in prediction~\cite{plummer2015flickr30k,fidler_grounding_graph, tellex2011approaching, hu2017learning}.
Our problem is different as we must reason over all possible graph structures. 
Deep sequential models have demonstrated strong performance for tasks such as captioning~\cite{larrycaption,msrcaption, googlecaption, stanfordcaption} and visual question answering~\cite{vqa1,vqa2,vqa3,vqa4,vqa5}, including for problems not traditionally not thought of as sequential, such as multilabel classification~\cite{cnnrnn16}. Indeed, graph linearization has worked surprisingly well for many problems in vision and language, such as generating image captions from object detections~\cite{obj2textEMNLP2017}, language parsing~\cite{vinyals2015grammar}, generating text from abstract meaning graphs~\cite{konstas2017neural}. Our work leverages the ability of RNNs to memorize long sequences in order to capture graph motifs in Visual Genome. Finally, recent works incorporate recurrent models into detection and segmentation~\cite{polygonrnn,ren2016end} and our methods contribute evidence that RNNs provide effective context for consecutive detection predictions.

\paragraph{Scene Graph Methods}
Several works have explored the role of priors by incorporating background language statistics~\cite{lu_visual_2016, Yu_2017_ICCV} or by attempting to preprocess scene graphs~\cite{zhang_learning_2016}.
Instead, we allow our model to directly learn to use scene graph priors effectively. 
Furthermore, recent graph-propagation methods were applied but converge quickly and bottle neck through edges, significantly limiting information exchange ~\cite{xu_scene_2017, li2017msdn, Dai2017DetectingVR, li2017vip}.
On the other hand, our method allows global exchange of information about context through conditioning and avoids uninformative edge predictions until the end.
Others have explored creating richer models between image regions, introducing new convolutional features and new objectives~\cite{DBLP:journals/corr/NewellD17,Zhang2017VisualTE, li2017msdn, liang_deep_2017}. Our work is complementary and instead focuses on the role of context. See the supplemental section for a comprehensive comparison to prior work. 

\section{Conclusion}
\label{sec:conclusion}
We presented an analysis of the Visual Genome dataset showing that motifs are prevalent, and hence important to model.
Motivated by this analysis, we introduced strong baselines 
that improve over prior state-of-the-art models by modeling these intra-graph interactions, while mostly ignoring visual cues.
We also introduced our model \model~for capturing higher order structure and global interactions in scene graphs that achieves additional significant gains over our already strong baselines.

\section*{Acknowledgements}
\label{sec:acknowledgements}
We thank the anonymous reviewers along with Ali Farhadi and Roozbeh Mottaghi for their helpful feedback. This work is supported by the National Science Foundation Graduate Research Fellowship (DGE-1256082), the NSF grant (IIS-1524371, 1703166), DARPA CwC program through ARO (W911NF-15-1-0543), IARPA's DIVA grant, and gifts by Google and Facebook.
\section*{Supplemental}
\label{sec:supp}
\begin{table*}[!t]
\setlength\tabcolsep{2.5pt} 
\small
\centering
\begin{tabular}{@{}l@{\hspace{0.2em}}l cc |cc |cc | cc |cc |cc |cc| cc @{}}
\toprule
& &\multicolumn{6}{c|}{Graph constraints} & \multicolumn{10}{c}{No graph constraints} \\ 
&& \multicolumn{2}{c|}{\textsc{SGDet}} & \multicolumn{2}{c|}{\textsc{SGCls}} & \multicolumn{2}{c|}{\textsc{PredCls}} & \multicolumn{2}{c|}{\textsc{SGDet}} & \multicolumn{2}{c|}{\textsc{SGCls}} & \multicolumn{2}{c|}{\textsc{PredCls}} &
\multicolumn{2}{c|}{\textsc{PhrDet}} & \multicolumn{2}{c}{\textsc{PredDet}} \\
& Model & \tiny{R$@$50}  & \tiny{R$@$100} & \tiny{R$@$50}  & \tiny{R$@$100}& \tiny{R$@$50}  & \tiny{R$@$100}& \tiny{R$@$50}  & \tiny{R$@$100}& \tiny{R$@$50}  & \tiny{R$@$100}& \tiny{R$@$50}  & \tiny{R$@$100}& \tiny{R$@$50}  & \tiny{R$@$100}& \tiny{R$@$50}  & \tiny{R$@$100} \\ \hline
\multirow{8}{*}{\rotatebox[origin=c]{90}{\cite{xu_scene_2017}'s split}} & \textsc{Vrd} \cite{lu_visual_2016}, from \cite{xu_scene_2017} & 0.3  & 0.5 & 11.8  & 14.1  & 27.9  & 35.0 &&&&&&&&&\\ 
&\sc Assoc. Embed 
\cite{DBLP:journals/corr/NewellD17} & 8.1 & 8.2 & 21.8 & 22.6 & 54.2 & 55.5 & 9.7 & 11.3 & 26.5 & 30.0 & 68.0 & 75.2&&&&\\ 
&\sc Message Passing \cite{xu_scene_2017}   & 3.4  & 4.2 &  21.7 & 24.4 & 44.8 & 53.0 &&&&&&&&&\\ 
&\sc Message Passing$+$ & 20.7 & 24.5 & 34.6 & 35.4 & 59.3 & 61.3 &  22.0 & 27.4 & 43.4 & 47.2 & 75.2 & 83.6 & 34.4 & 42.2 & 93.5 & 97.2 \\
&\textsc{Freq} & 23.5 & 27.6 & 32.4 & 34.0 & 59.9 & 64.1 &
25.3& 30.9& 40.5& 43.7& 71.3& 81.2&37.2&45.0& 88.3& 90.1 \\
&\textsc{Freq-Overlap} & 26.2 & 30.1 & 32.3 & 32.9 & 60.6 & 62.2 &
28.6& 34.4& 39.0& 43.4& 75.7& 82.9& 41.6& 49.9 & 94.6& 96.9\\ 
&\model-\textsc{NoContext} & 26.2 & 29.0 & 34.8 & 35.5 & 63.7 & 65.6 & 29.8 & 34.7 & 43.4 & 46.6 & 78.8 & 85.9 & 43.5 & 50.9 & 94.2 & 97.1 \\
&\model & {\bf 27.2} & {\bf 30.3} & {\bf 35.8} & {\bf 36.5} & {\bf 65.2} & {\bf 67.1} & {\bf 30.5} & {\bf 35.8} & {\bf 44.5} & {\bf 47.7} & {\bf 81.1} & {\bf 88.3} & {\bf 44.2} & {\bf 52.1} & {\bf 96.0} & {\bf 98.4} \\ \hline
\multirow{4}{*}{\rotatebox[origin=c]{90}{\cite{li2017msdn} split}} &\textsc{MSDN} \cite{li2017msdn}$\star$  & 10.7 & 14.2 & 24.3 & 26.5 & 67.0 & 71.0 &&&& &&&&& \\ 
&\textsc{MSDN}  & 11.7 & 14.0 & 20.9 & 24.0 & 42.3 & 48.2 &&&& &&&&& \\ 
&\textsc{MSDN}-\textsc{Freq} & {\bf 13.5} & {\bf 15.7} & {\bf 25.8} & {\bf 27.8} & {\bf 56.0} & {\bf 61.0} &&&& &&&&& \\ 
&\textsc{SCR}\cite{li2017vip} &&&&&&&10.67 & 13.81 &&&&&16.58 & 21.54&&  \\
\hline
\multirow{4}{*}{\rotatebox[origin=c]{90}{other split}} &\textsc{DR-Net}\cite{Dai2017DetectingVR} & &&&&&& 20.79 & 23.76 && & & & 23.95 & 27.57 & 88.26 & 91.26 \\
&\textsc{VRL}\cite{liang_deep_2017} &12.57 & 13.34&&&&&&& &&&&14.36 & 16.09 && \\
&\textsc{VTE}\cite{Zhang2017VisualTE}  &&&&&&& 5.52 & 6.04 &&&&& 9.46 & 10.45 && \\
&\textsc{LKD}\cite{Yu_2017_ICCV} &&&&&&& &  &&&&&&& 92.31  &  95.68 \\
\bottomrule
\end{tabular}
\caption{Results with and without scene graph constraints. Horizontal lines indicate different dataset preprocessing settings (the ``other split'' results, to the best of our knowledge, are reported on different splits). $\star$: \cite{li2017msdn} authors acknowledge that their paper results aren't reproducible for \textsc{SGCls} and \textsc{PredCls}; their current best reproducible numbers are one line below. \textsc{MSDN}-\textsc{Freq}: Results from using node prediction from \cite{li2017msdn} and edge prediction from \textsc{Freq}.
}\label{tab:supptable}
\end{table*}
Current work in scene graph parsing is largely inconsistent in terms of evaluation and experiments across papers are not completely comparable. 
In this supplementary material, we attempt to classify some of the differences and put the works together in the most comparable light. 
\subsection*{Setup}
In our paper, we compared against papers that (to the best of our knowledge) evaluated in the same way as \cite{xu_scene_2017}. 
Variation in evaluation consists of two types:
\begin{itemize}
    \item Custom data handling, such as creating paper-specific dataset splits, changing the data pre-processing, or using different label sets.
    \item Omitting graph constraints, namely, allowing a head-tail pair to have multiple edge labels in system output. We hypothesize that omitting graph constraints should always lead to higher numbers, since the model is then allowed multiple guesses for challenging objects and relations.
\end{itemize}
Table~\ref{tab:supptable} provides a best effort comprehensive review against all prior work that we are aware of. Other works also introduce slight variations in the tasks that are evaluated:\footnote{We use task names from \cite{lu_visual_2016}, despite inconsistency in whether the underlying task actually involves classification or detection.}

\begin{itemize}
    \item \textbf{Predicate Detection} (\textsc{PredDet}). The model is given a list of labeled boxes, as in predicate classification, and a list of head-tail pairs that have edges in the ground truth (the model makes no edge predictions for head-tail pairs not in the ground truth).
    \item \textbf{Phrase Detection} (\textsc{PhrDet}). The model must produce a set of objects and edges, as in scene graph detection.  An edge is counted as a match if the objects and predicate match the ground truth, with the IOU between the {\bf union-boxes} of the prediction and the ground truth over 0.5 (in contrast to scene graph detection where each object box must independently overlap with the corresponding ground truth box). 
\end{itemize}
\subsection*{Models considered}
In Table~\ref{tab:supptable}, we list the following additional methods:
\begin{itemize}
    \item \textsc{MSDN} \cite{li2017msdn}: This model is an extension of the message passing idea from \cite{xu_scene_2017}. In addition to using an RPN to propose boxes for objects, an additional RPN is used to propose regions for captioning. The caption generator is trained using an additional loss on the annotated regions from Visual Genome.
    \item \textsc{MSDN-Freq}: To benchmark the performance on \cite{li2017msdn}'s split (with more aggressive preprocessing than \cite{xu_scene_2017} and with small objects removed), we evaluated a version of our \textsc{Freq} baseline in \cite{li2017msdn}'s codebase. We took a checkpoint from the authors and replaced all edge predictions with predictions from the training set statistics from  \cite{li2017msdn}'s split. 
    \item \textsc{SCR} \cite{li2017vip}: This model uses an RPN to generate triplet proposals. Messages are then passed between the head, tail, and predicate for each triplet.
    \item \textsc{DR-Net} \cite{Dai2017DetectingVR}: Similar to \cite{xu_scene_2017}, this model uses an object detector to propose regions, and then messages are passed between relationship components using an approximation to CRF inference. 
    \item \textsc{VRL} \cite{liang_deep_2017}: This model constructs a scene graph incrementally. During training, a reinforcement learning loss is used to reward the model when it predicts correct components.
    \item \textsc{VTE} \cite{Zhang2017VisualTE}: This model learns subject, predicate, and object embeddings. A margin loss is used to reward the model for predicting correct triplets over incorrect ones. \item \textsc{LKD} \cite{Yu_2017_ICCV}: This model uses word vectors to regularize a CNN that predicts relationship triplets.
\end{itemize}

\subsection*{Summary}
The amount of variation in Table~\ref{tab:supptable} requires extremely cautious interpretation. 
As expected, removing graph constraints significantly increases reported performance and both predicate detection and phrase detection are significantly less challenging than predicate classification and scene graph detection, respectively.
On \cite{li2017msdn}'s split, the \textsc{MSDN-Freq} baseline outperforms \textsc{MSDN} on all evaluation settings, suggesting baseline is robust across alternative data settings.
In total, the results suggest that our model and baselines are at least competitive with other approaches on different configurations of the task.
\newpage

{\small
\bibliographystyle{ieee}
\bibliography{zoterobib}
}

\end{document}